# Recent, Rapid Advancement in Visual Question Answering: a Review

Venkat Kodali
*Department of Information Science*
*University of Arkansas at Little Rock*
Little Rock, USA
vxkodali@ualr.edu

Daniel Berleant
*Department of Information Science*
*University of Arkansas at Little Rock*
Little Rock, USA
jdberleant@ualr.edu

*Abstract*—Understanding visual question answering is going to be crucial for numerous human activities. However, it presents major challenges at the heart of the artificial intelligence endeavor. This paper presents an update on the rapid advancements in visual question answering using images that have occurred in the last couple of years. Tremendous growth in research on improving visual question answering system architecture has been published recently, showing the importance of multimodal architectures. Several points on the benefits of visual question answering are mentioned in the review paper as in [1], on which the present article builds, including subsequent updates in the field.

## I.   Introduction

Image understanding has been one of the primary drivers of artificial intelligence research and development over the years. In the healthcare sector, much of many physicians' practices depends on diagnostic radiology such as CT, CAT, PET, MRI, X-rays, etc. Pathology is widely used to make the initial diagnosis, but nevertheless medical images help in a major way for physicians to identify and treat health problems. For cancer tumors, lab results might not tell if the patient has a tumor in the lung or brain. Using radiology images, doctors can often much more clearly identify the presence and location of tumors in human body. Artificial intelligence is sometimes considered an exponential technology, and has advanced greatly in just a couple of decades, including speeding up tremendously in the last few years, due especially to advances in AI algorithms. This paper is written to provide an understanding of recent advancement in VQA. Newer techniques in both the image understanding and natural language understanding (NLU) subfields of AI have been helping to enable progress in visual question answering. The focus has been on connectionist approaches, in particular their recent and rapidly improving models. Thus, application of the deep learning approach to connectionism has been widely studied in the fields of speech recognition, drug design, precision medicine, disease detection, health surveillance, health care, image classification, etc.

## II.   The Visual Question Answering Concept

Publication of this work was supported by the National Science Foundation under Award No. OIA-1946391. The content reflects the views of the authors and not necessarily the NSF.

VQA involves both computer vision and natural language processing, exemplifying what is commonly referred to as multi-model tasks. In visual question answering, the model is trained to output a correct answer for an image and a related question. Fig. 1 shows an example of visual question answering.

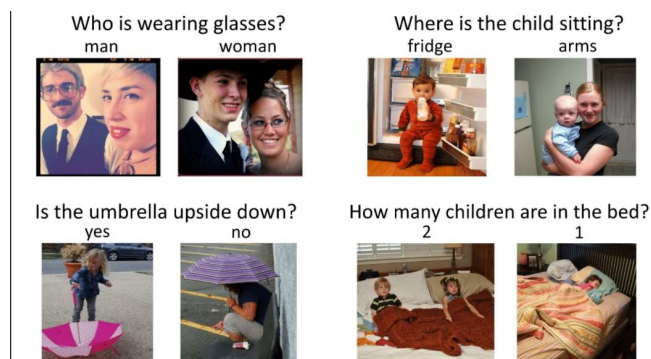

Fig. 1. Example picture of VQA on images (Source — VQA: Visual Question Answering, visualqa.org).

Fig. 2 shows a schematic of the VQA process. The objective of VQA is to predict the answer to a question about an image.

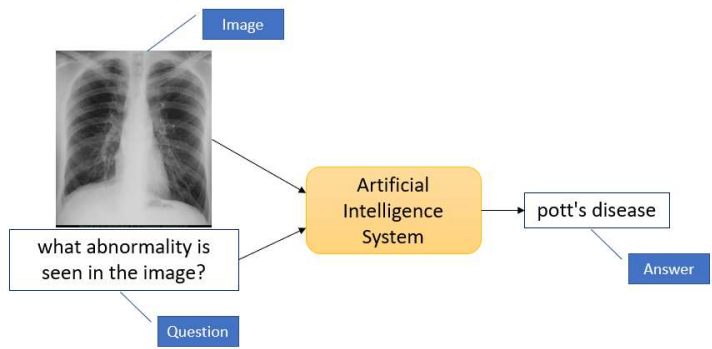

Fig. 2. Visual question answering process.

## III.   VQA papers/articles

Deep learning research has produced several new techniques in VQA. Fig. 3 illustrates that there were very few research papers/articles until 2015. The data for the graph is from a search query qualified by year in Google Scholar on the query "visual question answering." After 2017 the growth in the research work has increased dramatically. The number of research papers increased from 73 to 3,400 per year between year 2015 and 2021, more than a 40-fold increase in six years. Clearly, interest in VQA is currently in a period of explosive growth.

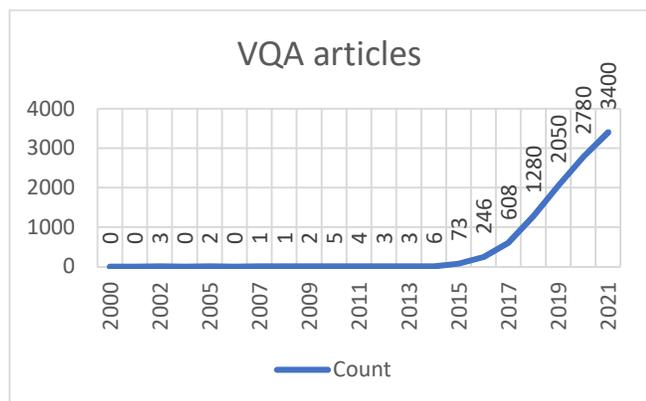

Fig.3. VQA articles per year (raw data source: Google Scholar)

The trend in recent years shows that there are numerous papers being published at an increasing rate dealing with VQA. Most of these papers involved non-medical images. VQA on medical images is challenging due to the often-low resolution nature of these images and the frequent complexity of interpreting medical images. Manmadhan and Kavoor [1] covered this aspect in more detail and their observations remain relevant today.

Although [1] has written about the state of art of VQA, the most recent article it cites is from June 2019 [26]. The present article addresses newer contributions, needed due to the rapid advancement in VQA since then including significant newer techniques that have been developed.

## IV. MEDICAL DATASETS

VQAMed (2021) provides medical images and question answer datasets. Fig. 4 illustrates the numbers of words in each answer from the 2019 VQAMed image dataset.

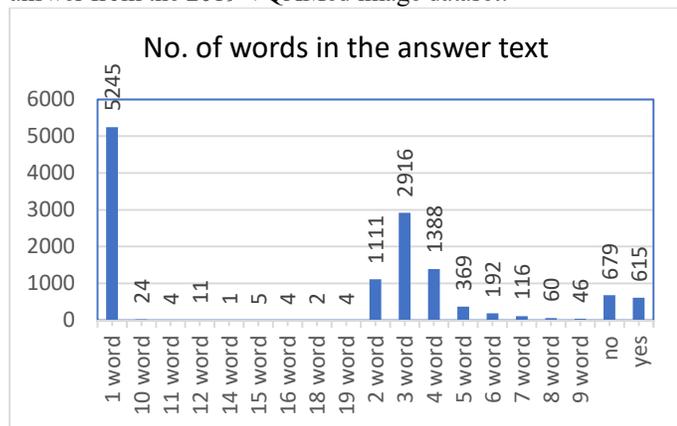

Fig. 4. VQAMed dataset from 2019 – word number breakdown.

The complexity of providing an accurate answer increases with the number of words in the answer.

## V. RESOURCES

Some medical image datasets available online are the following:

1) *OASIS Brains (Open Access Series of Imaging Studies).* This is a multi-modal neuroimaging dataset available at [32].
2) *The Cancer Imaging Archive (TCIA 2022).* TCIA is a publicly available archive of medical images of cancer. The images contain different modality images such as MRI, CT etc. The images are in DICOM format.
3) *The Cancer Genome Atlas Lung Adenocarcinoma (TCGA-LUAD).* This corpus provides clinical images [34].
4) *The SICAS Medical Image Repository.* SICAS [35] is large repository of medical images such as CT, microCT, etc.

Prominent relevant image models trained on very large medical image datasets include the following.

1) *ImageNet*: an image dataset containing millions of labeled and sorted images [27].
2) *CheXNet:* a 121-layer DenseNet trained on ChestX-ray14 for pneumonia detection [28].

TABLE I lists some of the VQA datasets.

TABLE I

| Year | VQA Dataset | |
|------|------|------|
| | Name | Description |
| 2014 | DAQUAR | DAtaset for QUestion Answering on Real-world images. |
| 2015 | COCO-QA. COCO | Large-scale object detection, segmentation, and captioning dataset. |
| 2016 | Visual7W | Large visual question answering (QA) datase. |
| 2018 | VQAMed | Medical dataset available in the ImageCLEFF website starting in 2018. |

## VI. EXTRACTING IMAGE FEATURES

The following are commonly used architectures for image feature extraction.

1) *DensetNet.* A DenseNet (short for Dense Convolutional Network) feed-forward connects every layer to all the other layers [17].
2) *VGG16/VGG19.* VGG is a convolutional neural network architecture [18].
3) *ResNet (Residual Network).* ResNet is based on a residual learning framework [19].

4) *AlexNet.* AlexNet is a deep convolution neural network that was the first CNN to use GPUs to improve performance [20].
5) *GoogleNet.* The primary focus of this convolution neural network is improving the utilization of computer resources [21].

Figs. 5 & 6 categorize architectures for VQA based on research papers found online. The data for the graph was from searches on Google Scholar using the search queries is on terms such as Densenet121, VGG16, VGG19, ResNet, AlexNet and GoogleNet.

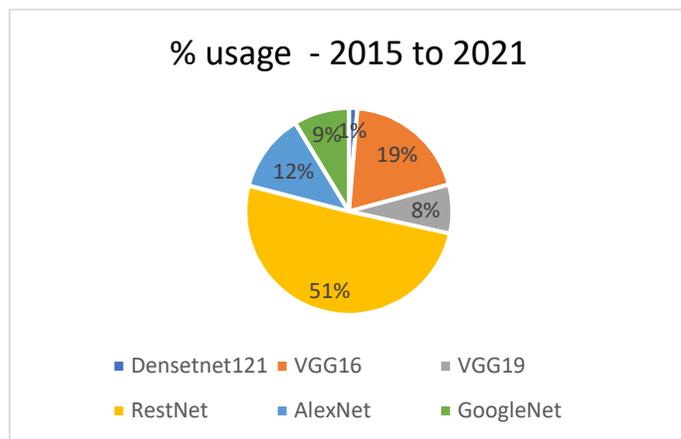

Fig. 5. Percentage usage of models. (Source: Google Scholar)

### VII. QUESTION EMBEDDING

One-hot vector encoding was one of the initial methods used in natural language processing for question embedding [25]. Bag of words (BOW) is a simple method for characterizing text data and is relatively straightforward to implement (Zhang et al. 2010 [13]). Word2Vec is accomplished using the skip-gram and continuous bag of words (CBOW) methods. The key aspect of the Word2Vec approach is to group similar words (Church 2017 [14]). Global Vectors (GloVe) is a unsupervised learning algorithm was designed to incorporate global word co-occurrence statistics in its vector encodings (Pennington et al. 2014 [15]). Embedding from language (ElMo) is a method for text data embedding. ELMo is a natural language processing framework developed by AllenNLP (Peters et al. 2018 [16]). ELMo word vectors are calculated using a two-layer bidirectional language model. In a bidirectional language model, word prediction is based on both the preceding and following words.

### VIII. EARLIER VQA RESEARCH

Manmadhan [1] covered numerous developments until June of 2019. We begin by supplementing that and also include mentions of early work. The year 2015 is foundational for VQA. Here are several VQA methods proposed since 2015.

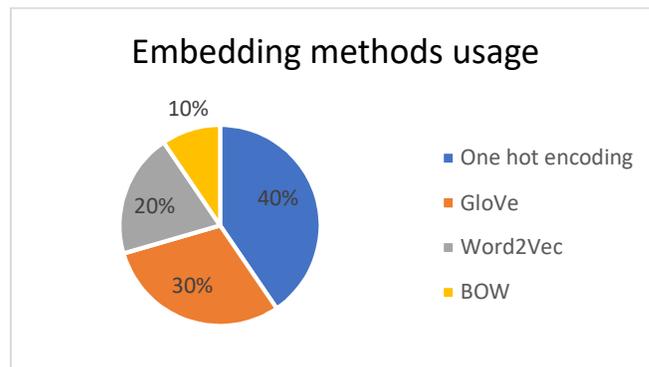

Fig. 6. Percentage usage of several word embedding methods. (Source: Google Scholar, search result is based on: "one hot encoding", GloVe, Word2Vec and BOW.)

Geman et al. (2015 [5]) addresses a visual Turing test to predict single answers such as yes/no for questions on images. The Turing test model however turns out to be limited in its applicability to VQA.

Antol (2015 [4]) introduced the concept of VQA on images using computer vision and natural language processing on free form and open-ended questions. Their research encompasses multi-modal and multi-discipline artificial intelligence. Goal driven tasks develop a model for a specific application, such as helping the visually impaired or monitoring video images for security. Free form text in this context is text that does not have a particular structure. Open-ended questions are characterized by lack of constraints on the type of answer expected, while open-ended answers are distinguished from multiple-choice answers by being much more unconstrained. The study includes a comparison between image captioning and image question answering.

### IX. ENCODER DECODER

The encoder decoder model was found to be an improved solution for certain natural language processing tasks, in particular for transfer learning from pretrained models.

Kafle and Kanan (2016 [8]) experimented with the encoder decoder model using recurrent neural network (RNN) with gated units on image datasets such as COCO (Common Object in COntext) and DAQUAR (DAtaset for QUestion Answering on Real-world images).

Later, the encoder decoder based Long Short-Term Memory (LSTM) model gained popularity and can provide better accuracy. Singh (2016 [9]) experimented with LTSM + CNN and showed that it outperformed other models tested in the paper.

### X. VQA APPROACHES

The following are some of the VQA approaches used in the early research on VQA:

1) *Stacked Attention Network (SAN):* SAN is for image question answering using a question semantic representation (Yang et al. 2016 [36]).
2) *Multi-Modal Factorized Bilinear Pooling (MFB):* Introduced by Fukui et al. (2016 [37]), MFB combines multi-modal features for visual question answering. Integrating MFB and co-attention learning, this architecture is compared with other bilinear pooling approaches.
3) *Answer Type Prediction*: Reference [8] introduced an approach to answer type prediction using a combination Bayesian framework and discriminative model.
4) *Fact Based VQA:* This model relies on excluding questions that require factual knowledge to answer (Wang et al. 2016 [38]).
5) *Attention Based VQA:* Chen et al. (2015 [39]) introduced the attention based configurable convolution neural network model.
6) *Focus Regions for VQA:* This method is based on learning select image regions relevant to the annotated question and answer (Shih et al. 2016 [40].
7) *A Focused Dynamic Attention Model (FDA) for VQA:* FDA (Ilievski et al. 2016 [41]) was proposed to better align representations of image content with the questions.
8) *Dual Attention Network for Visual Question Answering:* This work (Xu and Saenko 2016 [52]) applied attention to both images and question features for VQA, instead of just images.
9) *Structured Attention for VQA:* (Zhu et al. (2017 [54]) proposed a visual attention model based on a multivariate distribution over a grid structured conditional random field on image regions.
10) *Graph-Structured Representation of VQA:* Teney et al. 2017 [55]) introduced a graph-structured representation model based on building graphs over scene objects and question words, processing them using the structure of these representations.
11) *Feature Embedding for Visual Question Answering:* This approach (Lu et al. 2018 [62]) combined visual attention on free-form regions and multiplicative feature embedding.
12) *Semantically Guided VQA:* This approach (Zhao et al. 2018 [63]) applied multiple instances learning to extract visual representations, including entities such as actions and colors.
13) *Deep Modular Co-Attention Networks for VQA:* This model (Yu et al. 2019 [42]) uses deep modular co-attention layers. Each layer supports question-guided attention to images using a modular composition of two basic attention units.

A. *VQA approaches in 2019*

There was great progress in VQA research in 2019 and most of that research involved transformers. Key to 2019's advances was improvements in Bidirectional Encoder Representations from Transformers (BERT) for natural language processing. Researchers introduced unique methods for applying this approach to VQA. Innovative new methods such as co-attention transformers, large-scale transformer models that consists of three encoders, the cross-modality model with self-attention and cross-attention layers, and modification of the orginal BERT by assigning new elements such visuals defined by regions of interest in images and linguistic elements as inputs. Identifying regions of interest can help algorithms speed up object detection. The attention concept continued to bear fruit, such as with self-attention within a transformer, and encoding images into graphs to model inter-object relations using a graph attention mechanism.

The following are some of the most influential relevant articles of the year.

1) *ViLBERT: Pretraining Task-Agnostic Visiolinguistic Representations for Vision-and-Language Tasks:* Lu et al. [43] introduced VilBERT "for learning task-agnostic joint representations of image content and natural language."
2) *LXMERT: Learning Cross-Modality Encoder Representations from Transformers:* Tan and Bansal [44] introduced LXMERT to "learn vision and language associations with a transformer model consisting of three encoders."
3) *Vl-bert: Pre-Training of Generic Visual-Linguistic Representations:* Su et al. [45] introduced generic representations for linguistic tasks adopting a transformer model. The foundation of VL-BERT is a "transformer attention module taking both visual and linguistic embedded features as input."
4) *Relation-Aware Graph Attention Network for Visual Question Answering.* Li et al. [47] introduced this "relation aware graph attention network which encodes each image into graph and models multi-type inter object relation via a graph attention mechanism."

B. *VQA approaches in 2020*

In 2020, a variety of new models for VQA were introduced. Influential articles described models using methods such as grid-based convolutional features for VQA, bottom-up top-down models and ensemble-based models. Also used were simple CNN+LSTM models, an attention-based model strategy and a compositional model approach. A new method introduced multi-objective visual relationship detection. Another interesting concept that was introduced is mutations of inputs (questions and images).

The following are some of most visible articles published in the year 2020.

1) *In Defense of Grid Features for Visual Question Answering:* Jiang et al. [48] revisit grid-based convolution features for VQA.

2) *Counterfactual Samples Synthesizing for Robust Visual Question Answering:* Chen et al. [49] describe a framework to address "visually explainability (the model should rely on the right visual regions) and question sensitivity (the model should be sensitive to linguistic variation in questions)."
3) *Towards Causal VQA: Revealing and Reducing Spurious Correlations by Invariant and Covariant Semantic Editing.* Agarwal et al. [50] introduced causal VQA. The method performs automated semantic image manipulations and tests.
4) *Visual Question Answering Model Based on Visual Relationship Detection:* Xi et al. [51] introduce model based multi-objective visual relationship detection. What they call appearance features replace image features from the original object.
5) *Overview of the VQA-Med Task at ImageCLEF 2020: Visual Question Answering and Generation in the Medical Domain:* Yu et al. [64] provide a well-regarded survey of the submitted works on visual question answering on medical images.
6) *Training Paradigm for Out-of-Distribution Generalization in Visual Question Answering:* Gokhale et al. [65] introduce a model using consistency constraints on the training goal. This permits conclusions about the effects of meaning changes in the questions.

### C. VQA approaches in 2021

In year 2021, a wide range of articles were published and the number of articles increased significantly. Research in some of the popular articles addressed language bias, large VQA datasets, the tensor decomposition model, dynamic word vectors, text aware pretraining, bias evaluating metrics, and others. All the articles demonstrated improvements in solutions for VQA. The following are some of these articles.

1) *Counterfactual VQA: a Cause-Effect Look at Language Bias:* Niu et al. [66] introduced a framework to "capture […] language bias as the direct causal effect of questions on answers and reduce the language bias by subtracting the direct language effect from the total cause." This novel method is important since training on biased text will impact predictions.
2) *DocVQA: A Dataset for VQA on Document Images:* Mathew et al. [71] provided a corpus of 12,767 varied images associated with 50,000 questions and answers. The authors introduced DocVQA "as a high-level task dynamically driving DAR algorithms to conditionally interpret document images." The underlying concept is to include all other contents in a document besides regular text.
3) *DecomVQANet: Decomposing Visual Question Answering Deep Network Via Tensor Decomposition and Regression:* Bai et al. [67] introduced DecomVQANet "to conduct various decomposition methods and regression strategies on different layers, including Canonical Polyadic, Tucker, and Tensor Train to decompose fully connected layers in CNN and LSTM" networks.
4) *Joint Embedding VQA Model Based on Dynamic Word Vector:* Ma et al. [68] introduced this approach, concluding that dynamic word vectors outperform static word vectors on the VQA task.
5) *TAP: Text-Aware Pre-Training for Text-VQA and Text-Caption:* Yang et al. introduced TAP, which uses scene text in a pretraining model and later fine tunes to improve it.
6) *Roses Are Red, Violets Are Blue... but Should VQA Expect Them To*? Kervadec et al. [70] proposed "a new fine-grained re-organization of the GQA dataset and a set of the respective evaluation metrics allowing to precisely evaluate the reasoning behavior of VQA models."

## XI. TRANSFORMERS

Transformers (Viswani et al. 2017 [10]) improve upon some of the previous state of art solutions such as recurrent neural networks (RNN), long short-term memory (LSTM).

### A. BERT : Bidirectional Encoder Representations from Transformers

BERT (Devlin et al. 2018 [24]) builds on transformers. It has been used extensively for various natural language processing tasks. While encoder decoder models have sequential layers, in which basically inputs are sequences and outputs are a single number, BERT models are transformer based. BERT models are of two types: BERT base and BERT large. BERT base using 12 encoder layers and BERT large using 24 encoder layers.

### B. Vision Transformer

Research has shown the benefits of using transformers in natural language processing. The concept has also been applied to computer vision, resulting in the vision transformer (ViT) model.

BERT's success has opened up research directions focused on applying similar techniques on images. In the vision transformer model, the images are divided into patches, each patch with size 16x16 pixels. Dosovitskiy et al. (2021 [7]) introduce vision transformers, finding that images divided into a sequence of patches and processed by a transformer performed better for image classification tasks. Fig. 7 gives an example of creating image patches from a single image.

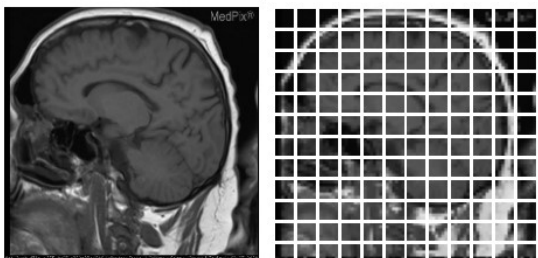

Fig. 5. Image patches

*C. Swin Transformers*

Swin transformers were introduced (Liu et al. 2016 [6]) to overcome challenges in adapting transformers for computer vision. Swin transformers applied to ImageNet had higher accuracy compared to vision transformers, which themselves are improved over CNNs.

## XII. CONCLUSIONS

We are continuing to see a vigorous upward trend in development of VQA systems since the trend became evident in 2015. The state of art in VQA is changing rapidly each year with multiple improved architectures being proposed. In the year 2015, it was the deep neural networks models that were the state of art. They have given way to the much more efficient transformer architectures, which are themselves experiencing rapid Vita progress. According to Lu (2020 [2]), artificial intelligence is progressing rapidly in a way that the massive need for computing resource is likely to bring back Moore's Law back to life. The newer methods such a BERT and ViT have been found to be fast to train compared to more traditional neural networks.

Replicating results has been found to often be difficult or impossible in science (Haelle 2021 [3]) and there are a number of different reasons why this might be true for a given report. These include that the architecture might not be sufficiently explained, all the steps might not be defined, etc. Although this problem is unlikely to be resolved in the near future for science in general, and can only impeded progress according to standard views of scientific progress, it is clear that AI and, specifically, VQA have nevertheless been undergoing rapid advancements in recent years and there is little indication at present of any upcoming limit to continued progress.